\documentclass[10pt,twocolumn,letterpaper]{article}

\usepackage[pagenumbers]{cvpr} 

\usepackage{graphicx, amsmath, amssymb, caption, subcaption, multirow, overpic, textpos}
\usepackage[table]{xcolor}
\usepackage{makecell}
\usepackage[english]{babel}
\usepackage[pagebackref=false, breaklinks=true, colorlinks,citecolor=citecolor, linkcolor=linkcolor, bookmarks=false]{hyperref}
\definecolor{citecolor}{HTML}{0071BC}
\definecolor{linkcolor}{HTML}{ED1C24}

\def\cvprPaperID{**}
\def\confName{****}
\def\confYear{****}

\newlength\savewidth\newcommand\shline{\noalign{\global\savewidth\arrayrulewidth
  \global\arrayrulewidth 1pt}\hline\noalign{\global\arrayrulewidth\savewidth}}
\newcommand{\tablestyle}[2]{\setlength{\tabcolsep}{#1}\renewcommand{\arraystretch}{#2}\centering\footnotesize}
\renewcommand{\paragraph}[1]{\vspace{1.25mm}\noindent\textbf{#1}}

\newcolumntype{x}[1]{>{\centering\arraybackslash}p{#1pt}}
\newcolumntype{y}[1]{>{\raggedright\arraybackslash}p{#1pt}}
\newcolumntype{z}[1]{>{\raggedleft\arraybackslash}p{#1pt}}
\usepackage{pifont}
\newcommand{\cmark}{\ding{51}}%
\newcommand{\xmark}{\ding{55}}%
\newcommand{\app}{\raise.17ex\hbox{$\scriptstyle\sim$}}

\newcommand{\x}{{\times}}
\definecolor{deemph}{gray}{0.6}

\definecolor{baselinecolor}{gray}{.9}
\newcommand{\baseline}[1]{\cellcolor{baselinecolor}{#1}}

\usepackage{booktabs}
\usepackage{algorithm, algorithmic}
\usepackage[toc,page]{appendix}
\usepackage{float}
\definecolor{mgreen}{RGB}{57, 181, 74}

\newfloat{figtab}{htb}{fgtb}
\makeatletter
  \newcommand\figcaption{\def\@captype{figure}\caption}
  \newcommand\tabcaption{\def\@captype{table}\caption}
\makeatother

\def \alambic {\includegraphics[width=0.02\linewidth]{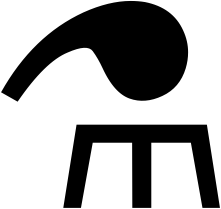}\xspace}

\usepackage[capitalize]{cleveref}
\Crefname{section}{Sec.}{Secs.}
\Crefname{section}{Section}{Sections}
\Crefname{table}{Table}{Tables}
\Crefname{table}{Tab.}{Tabs.}

\def\ourmethod{{dBOT}\xspace}
\def\ourmethoddis{dBOT\alambic\xspace}
\def\ourmethodn{{dBOT}}

\usepackage{amsmath,amsfonts,bm}

\def\eqref#1{equation~\ref{#1}}

\def\1{\bm{1}}

\def\vtheta{{\bm{\theta}}}

\DeclareMathAlphabet{\mathsfit}{\encodingdefault}{\sfdefault}{m}{sl}
\SetMathAlphabet{\mathsfit}{bold}{\encodingdefault}{\sfdefault}{bx}{n}

\newcommand*{\affaddr}[1]{#1} 

\begin{document}
\title{
Exploring Target Representations for Masked Autoencoders
}

\author{%
	Xingbin Liu$^{1,2}$\footnotemark[1] 
	\quad Jinghao Zhou$^{2}$\footnotemark[1]
	\quad Tao Kong$^{2}$\footnotemark[2]
	\quad Xianming Lin$^{1}$ 
	\quad  Rongrong Ji$^{1}$\\
	\affaddr{$^{1}$Xiamen University} \quad \affaddr{$^{2}$ByteDance} \quad \\
}

\maketitle
\renewcommand{\thefootnote}{\fnsymbol{footnote}}
{\let\thefootnote\relax\footnotetext{
$^*$Equal contribution. $^\dagger$Corresponding author. \\
\indent\hspace{3mm}Work done during Xingbin's internship at ByteDance.}}

\begin{abstract}

Masked autoencoders have become popular training paradigms for self-supervised visual representation learning. These models randomly mask a portion of the input and reconstruct the masked portion according to assigned target representations. 
In this paper, we show that a careful choice of the target representation is unnecessary for learning good visual representation since different targets tend to derive similarly behaved models. 
Driven by this observation, we propose a multi-stage masked distillation pipeline and use a randomly initialized model as the teacher, enabling us to effectively train high-capacity models without any effort to carefully design the target representation. 
On various downstream tasks of classification, transfer learning, object detection, and semantic segmentation, the proposed method to perform masked knowledge \textbf{d}istillation with \textbf{bo}otstrapped \textbf{t}eachers (\textbf{\ourmethod}) outperforms previous self-supervised methods by nontrivial margins. 
We hope our findings, as well as the proposed method, could motivate people to rethink the roles of target representations in pre-training masked autoencoders.
The code and pre-trained models are publicly available at \url{https://github.com/liuxingbin/dbot}.
\end{abstract}

\section{Introduction}

\textbf{M}asked \textbf{I}mage \textbf{M}odeling (MIM)~\cite{mae,maskfeat, baevski2022data2vec, zhou2021ibot} has recently become an active research topic in the field of visual representation learning and establishes strong performance for vision recognition tasks, \eg, image classification, object detection, and semantic segmentation, which also surpasses traditional supervised learning~\cite{deit} mechanism.
To be specific, MIM randomly masks a portion of the input and then reconstructs the masked portion according to the transformed target, formulated as
\begin{equation}
\mathop{\mathrm{min}}_{\theta} \mathop{\mathbb{E}}_{x\sim \mathcal{D}} \mathcal{M}(\mathcal{T} (x \odot (1-M)), f_\theta(x \odot M)),
\label{eq:mim}
\end{equation}
where ``$\odot$'' means element-wise product; $M$ is the \emph{patch mask}; ``$x\odot M$'' represents ``unmasked patches'' and vice versa; 
$f_\theta(\cdot)$ is the learnable network to be pre-trained; 
$\mathcal{T}$ is the transformation function generating the reconstructed target. $\mathcal{T}$ can either be a parameterized network or a traditional image feature transformation method;
$\mathcal{M}(\cdot,\cdot)$ is the similarity measurement, \eg, $l2$-distance~\cite{mae}. A masked image passed through the network $f_\theta(x \odot M)$ to reconstruct the visual representation of the intact image with transformation $\mathcal{T}(x \odot (1-M))$. 

A crucial problem of MIM is how to choose the reconstructed target, \ie, $\mathcal{T}(\cdot)$ in \cref{eq:mim}.
Previous methods use disparate teacher networks to generate the reconstruction target.
BEiT~\cite{beit} employs a pre-trained DALL-E~\cite{dall-e} as the teacher network. 
In MaskFeat~\cite{maskfeat}, authors use HOG~\cite{hog}, MoCo~\cite{moco} and DINO~\cite{dino} features to perform MIM; 
MVP~\cite{wei2022mvp} employs a multi-modality model, CLIP~\cite{clip}, which is pre-trained by rich image-text pairs. 
MAE~\cite{mae} uses image pixels as the target, which functions likewise to a randomly initialized teacher network, as demonstrated in~\cref{ap:pixel}.
iBOT~\cite{zhou2021ibot} and data2vec~\cite{baevski2022data2vec} use the exponential moving average (EMA) strategy to update teacher's parameters $\phi$.
Though different methods differ in their architectural designs and optimization, the choice of the teacher network lies crucial for each method and calls for a systematic study.
In this work, we paraphrase a term \textbf{M}asked \textbf{K}nowledge \textbf{D}istillation (MKD) to focus our discussion on a special case of MIM where the target is generated by a parameterized network (teacher network), \ie, $\mathcal{T}(\cdot)=h_\phi(\cdot)$.
In this setting, $\mathbf{T}$ is the teacher network, and $f$ is the student network.

The purpose of our work is to investigate \textit{whether a careful design of the teacher network for MKD matters}.
Such exploration is nontrivial given that different teacher networks contain different knowledge we endued into the teacher network,
which may induce diverse behaviors for the student networks.
And the painstaking selection of the target representations in the field of MIM.
To this end, we compare student networks distilled by four teacher networks with different computation pipelines, \ie, DINO~\cite{dino} for contrastive learning, MAE~\cite{mae} for masked autoencoding, DeiT~\cite{deit} for supervised learning, and DALL-E~\cite{dall-e} for autoregressive generation.
Four teachers are all pre-trained on ImageNet-1K for a fair comparison.
To our surprise, although the behaviors of the teacher networks are very different, the distilled student networks share similar characters after several stages of masked knowledge distillation:
\textbf{(i)} the performance variance between student networks distilled from different teachers rapidly decreases.
\textbf{(ii)} the model weights and output features across layers within the networks share similar properties.

Such observations indicate that the design of target representation is not essential for learning good visual representations when pre-trained with multi-stage, \ie, \textit{teacher networks do not matter with multi-stage masked knowledge distillation.}
Exceptionally, we use a randomly initialized model as teacher to perform multi-stage masked knowledge distillation, and find that it performs as well as those initialized by pre-trained models with the exact same settings!
Using a \textit{random model} as teachers not only avoids an extra pre-training stage, but also alleviates the painstaking selection of the target representations.

Based on the above studies and observations, we naturally propose to perform masked knowledge \textbf{d}istillation with \textbf{bo}otstrapped \textbf{t}eachers, short as \textbf{\ourmethod} \includegraphics[width=0.3cm,height=0.3cm]{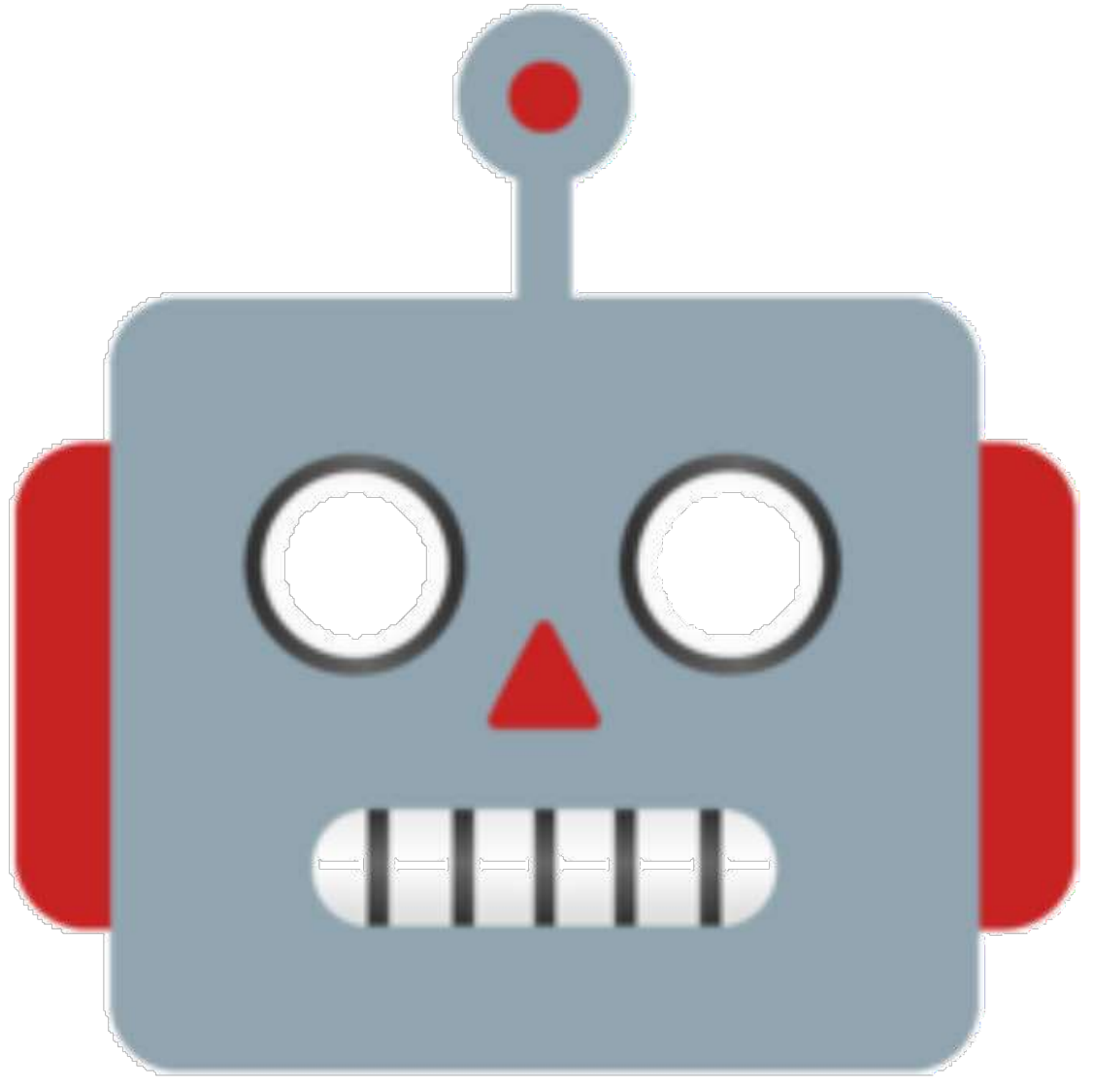}. 
Specifically, masked knowledge distillation is performed repeatedly in multiple stages. At the end of each stage, we assign the student's weight to the teacher and re-initialize the student's weight to continue masked knowledge distillation.
With simple yet effective design that enables pre-training starting from randomly initialized teachers, \ourmethod achieves 84.5\%, 86.6\%, and \textbf{88.0}\% top-1 fine-tuning accuracy on ImageNet-1K~\cite{imagenet} with ViT-B/16, ViT-L/16, and ViT-H/14, respectively, significantly surpassing previous states of the art, MAE.
Beyond that, \ourmethod achieves 52.7 and \textbf{56.0} AP$^\mathrm{box}$ for object detection on COCO~\cite{coco}, as well as 49.5 and \textbf{54.5} mIoU for semantic segmentation on ADE20K~\cite{ade}, with ViT-B/16 and ViT-L/16 respectively.
We also explore MKD with teachers of larger sizes, further boosting model performances on various visual tasks.

\section{Related work}
\subsection{Self-Supervised Visual Learning}
Self-supervised learning is an active research topic recently. 
Early practices revolve around contrastive learning~\cite{moco,simclr,byol,swav,dino} where 
the model output features of images transformed by different data augmentations are pulled together.
With the development of Masked Language Modeling (MLM) in language pre-training~\cite{bert}, researchers also introduce the training strategy of masked reconstruction to visual pre-training.
BEiT~\cite{beit} uses the DALL-E~\cite{dall-e} to encode an image patch as the target for model reconstruction. 
iBOT~\cite{zhou2021ibot} uses an online teacher shifting the target from offline to online to make the target semantic meaningful. 
In addition to using the token obtained from offline or online model as reconstruct target, MAE~\cite{mae}, SimMIM~\cite{xie2022simmim}, and MaskFeat~\cite{maskfeat} achieve good performance in masked-image reconstruction using low-level pixels or HOG~\cite{hog} features.
Among them, MAE uses an asymmetric encoder-decoder structure greatly increasing the training efficiency.
data2vec~\cite{baevski2022data2vec} demonstrates good generalizations on three modalities (vision, speech, and language) by reconstructing multiple neural network layer representations.

\subsection{Knowledge Distillation}
Knowledge distillation (KD) is widely employed in model knowledge compression~\cite{kd}, which improves the performance of the smaller student model by distilling the knowledge learned from a well-trained large teacher network. Further study on \eg relational KD~\cite{rkd}, contrastive KD~\cite{contrastivekd}, and latent feature KD~\cite{fitnets} is conducted to improve the performance of vanilla KD.
Beyond its prominence in the field of supervised learning, KD recently cuts a figure in self-supervised learning. 
Concurrent work manages to adopt conventional feature distillation~\cite{fd2022} to match contrastive models with MIM-trained ones. Nevertheless, it shows negligible gains on MIM-trained models such as MAE.
BEiT~\cite{beit}, MaskFeat~\cite{maskfeat} and MVP~\cite{wei2022mvp} could be seen as distilling knowledge from dVAE~\cite{dall-e}, HOG features~\cite{hog} and language-induced model CLIP~\cite{clip} within the discourse of MKD, respectively. 
Until now, there exists no work conferring a system-level study on the importance of how to choose adequate target representation or teacher networks to guide the learning of MKD.
Beyond that, we propose using a randomly initialized model as the teacher and bootstraps the teacher for stages, demonstrating superiority over other practices.

\begin{table*}
    \begin{center}
    \setlength{\tabcolsep}{1.45mm}
    \begin{tabular}{l|l|cccc|ccccc|ccccc}
        \multirow{2}{*}{\makecell[l]{computation\\ pipeline}}&\multirow{2}{*}{\makecell[l]{initialized\\ teacher}}& \multicolumn{4}{c|}{classification}& \multicolumn{5}{c|}{object detetion} & \multicolumn{5}{c}{semantic segmentation}\\
        &&0$^\mathrm{th}$  &1$^\mathrm{st}$ &2$^\mathrm{nd}$ &3$^\mathrm{rd}$ &0$^\mathrm{th}$ &1$^\mathrm{st}$ &2$^\mathrm{nd}$ &3$^\mathrm{rd}$ &4$^\mathrm{th}$ &0$^\mathrm{th}$ &1$^\mathrm{st}$ &2$^\mathrm{nd}$ &3$^\mathrm{rd}$ &4$^\mathrm{th}$ \\
        \hline
        Supervised & DeiT~\cite{deit} &   81.8&   83.6& \baseline{84.3}&   84.3&   49.1&   50.5& \baseline{52.5}&   52.4& - & 46.4&  49.2&  \baseline{50.4}&   49.9& -\\
        Contrastive & DINO~\cite{dino} &   83.2&   84.2& \baseline{84.5}&   84.4&   50.1&   52.5&  \baseline{52.9}&   52.7&-&   46.8& 49.7& \baseline{50.4} &   49.4 & -\\
        Autoregressive & DALL-E~\cite{dall-e} & 81.1 & 83.5 & \baseline{84.4} & 84.3 & 31.9 & 51.0 & \baseline{52.7} & 52.5 & - & 31.9 & 47.4 & \baseline{49.6} & 49.3 & -\\
        Autoencoding & MAE~\cite{mae} &    83.6&   84.3& \baseline{84.4}&   84.3&   50.6&   \baseline{52.9} &  52.7&   52.5& - & 48.1&   49.6& \baseline{50.4}&   49.8 & -\\
        -  & random& 77.3&   83.4& \baseline{84.5} &   84.3&   29.2&   49.6&   52.4& \baseline{52.7} & 52.4 & 25.7&   47.0&   49.1& \baseline{49.5} & 49.5\\
        \hline
        \multicolumn{2}{l|}{performance variance}&2.24 & 0.37 & 0.07 & 0.04 & 9.54 & 1.23  &  0.17 & 0.12 & - & 9.19 & 1.15 & 0.54 & 0.23 & -  \\
    \end{tabular}
    \end{center}
    \caption{The top-1 classification accuracy on ImageNet-1K, object detection AP-box on COCO with Cascade Mask R-CNN, and semantic segmentation mIoU on ADE20K with UperNet of \ourmethod using different models as the initialized teacher network. 
    Note that all models are pre-trained on ImageNet-1K, including DALL-E, for a fair comparison.
    We perform distillation in each stage for 800 epochs. In the 1$^\mathrm{st}$ stage, we distill from initialized teacher to obtain a student. In the subsequent (\ie, 2$^\mathrm{nd}$, 3$^\mathrm{rd}$, etc.) stages, the obtained students are leveraged as bootstrapped teacher to distill a new student.
    }
    \label{tab:different teacher}
\vspace{-0.2cm}
\end{table*}

\section{Does $h_\phi(\cdot)$ Matter in MKD?}
\label{sec:study}

Given the general form of masked knowledge distillation as shown in~\cref{eq:mim}, in this section, we aim to investigate \textit{whether the careful design of the target, \ie, teacher network $h_\phi(\cdot)$, matters}.  Specifically, we want to answer three questions as follows: 
\begin{itemize}
    \item Whether models distilled from different $h_\phi(\cdot)$ differ in terms of their transfer performances?
    \item Whether distilled models differ in terms of their weights and outputs?
    \item If $h_\phi(\cdot)$ does not matter, what matters more to close the gap between students distilled from different $h_\phi(\cdot)$?
\end{itemize}
To answer these questions, we employ the standard masked autoencoder framework~\cite{mae} to give a system-level study, introduced next.

\paragraph{Common setup.}
\label{study:setup}
The architectural settings strictly follow ~\cite{mae}.
For the teacher network, we use the vanilla ViT~\cite{vit} with intact input.
For the student network with masked input, we use the asymmetric encoder-decoder structure.
The student's output is further projected to a dimension the same as that of teacher's embedding. 
During pre-training, we use Smooth L1 loss~\cite{fast-rcnn} for the optimization of the student network, and the teacher network is kept fixed. Detailed settings are delayed to \cref{app:pretrain}.
We pre-train models on ImageNet-1K~\cite{imagenet} and conduct evaluation under classification on ImageNet, object detection on COCO~\cite{coco}, and semantic segmentation on ADE20K~\cite{ade}.

\subsection{Preliminary Study}
We first investigate the effect of using networks initialized differently as teachers for masked knowledge distillation.
Four canonical methods as \textbf{\textit{pre-trained teachers}} are substantiated, each from a category distinguished based on their computation pipelines, \ie, DeiT~\cite{deit} for supervised learning, DINO~\cite{dino} for contrastive learning, DALL-E~\cite{dall-e} for autoregressive generation, and MAE~\cite{mae} for autoencoding.
The results of initialized teacher at the 0$^\mathrm{th}$ stage and of its distilled student at the 1$^\mathrm{st}$ stage are shown in \cref{tab:different teacher}.

\paragraph{Different $h_\phi(\cdot)$ lead to similarly performed students.}
After the first stage of masked knowledge distillation, the student consistently outperforms teacher as shown in \cref{tab:different teacher}, yielding 1.8\%, 1.0\%, 2.4\%, and 0.7\% performance gains for four different $h_\phi(\cdot)$ respectively, demonstrating the effectiveness of masked knowledge distillation for visual representation learning. 
Although the performance order of different $h_\phi(\cdot)$ is reserved after the first stage of distillation, the students distilled from different $h_\phi(\cdot)$ have closer downstream performances compared to the original $h_\phi(\cdot)$. 
The performance variance drops from 2.24 to 0.37 after the first stage of distillation.
Take MAE and DALL-E being the initialized teachers as an example, the performance range in classification, drops from 2.5\% (83.6\% \textit{vs.} 81.1\%) to 0.8\% (84.3\% \textit{vs.} 83.5\%), indicating that the performance gap is narrowed after the first stage.
The conclusion holds true for experiments on object detection and semantic segmentation.

\subsection{Distillation with Multiple Stages}
Given the observations that better teacher generally induces better outperforming student, we are motivated to use the trained student as teacher to train new student repeatedly and study whether similar trend endures. If so, we would like to seek at what stage the performances saturate for different downstream tasks, as well as the discrepancy among the results incurred by different initialized teachers.

\paragraph{$h_\phi(\cdot)$ does not matter with multi-stage distillation.}
The performance gain is valid but decreases with multi-stage and eventually vanishes. Take MAE being the initialized teacher as an example, students outperform teachers by +0.7\%, +0.1\%, -0.1\% for classification, +2.3, -0.2, -0.2 points for object detection, and +1.5, +0.8, -0.6 points, for semantic segmentation, from the 0$^\mathrm{th}$ to the 3$^\mathrm{rd}$ stage. Other teachers and downstream tasks share the same conclusion.
Moreover, the performance gaps of students learned from different teachers decrease, especially after multi-stage, as shown by the performance variance at different stages in the last row of \cref{tab:different teacher}. Take classification tasks for instance, the variance decreases along with the training stage, \ie, 2.24, 0.37, 0.07, 0.04, which reveals that the choice of $h_\phi(\cdot)$ exerts little influence on the downstream performance. See \cref{tab:different teacher} for results of more downstream tasks.
To demonstrate models' differences in terms of weights and outputs, we conduct a property analysis in \cref{sec:analysis}.
Similar properties are found, which verify our conclusion.

\begin{figure*}[ht]
     \begin{center}
     \begin{subfigure}[t]{0.24\linewidth}
         \begin{center}
         \includegraphics[width=\linewidth]{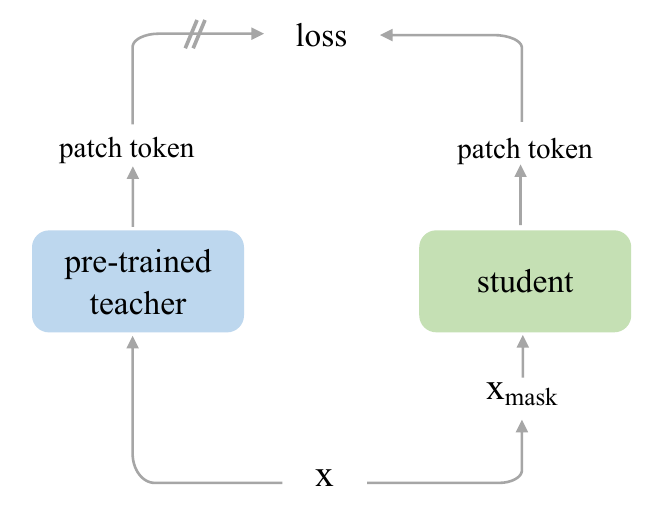}
         \end{center}
         \caption{BEiT~\cite{beit}}
         \label{fig:pre-trainedteacher}
     \end{subfigure}
     \hfill
     \begin{subfigure}[t]{0.24\linewidth}
         \begin{center}
         \includegraphics[width=\linewidth]{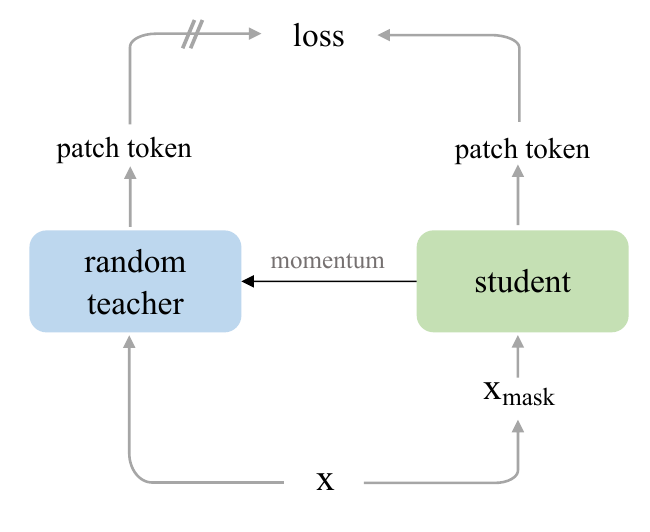}
         \end{center}
         \caption{iBOT~\cite{zhou2021ibot}, data2vec~\cite{baevski2022data2vec}}
         \label{fig:randomteacher}
     \end{subfigure}
     \hfill
     \begin{subfigure}[t]{0.5\linewidth}
         \begin{center}
         \includegraphics[width=\linewidth]{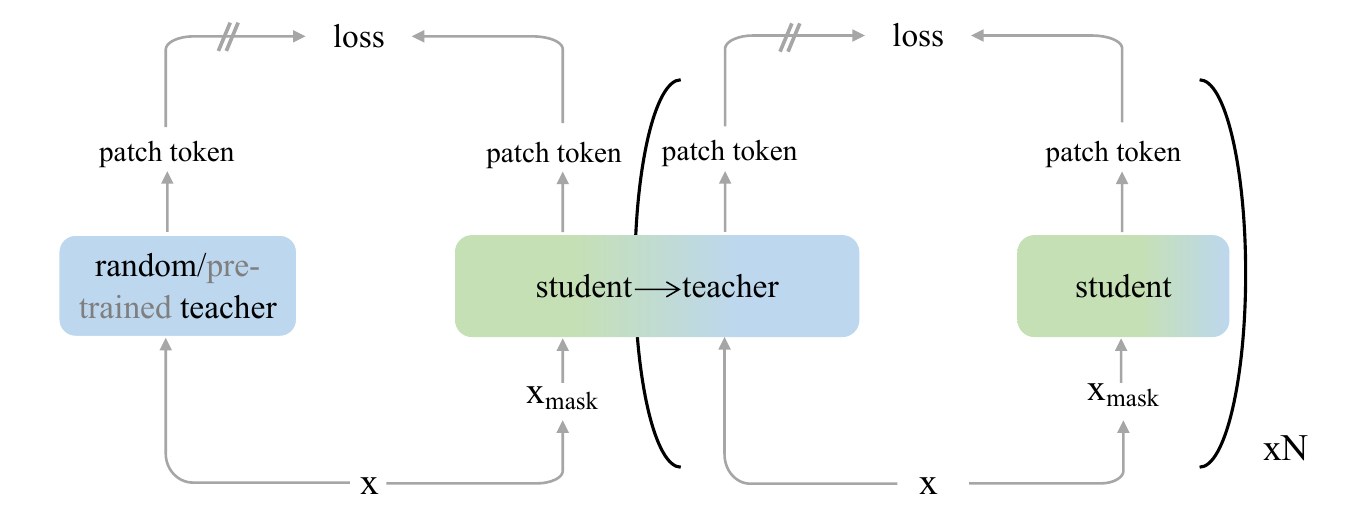}
         \end{center}
         \caption{\ourmethod}
         \label{fig:dbot}
   \end{subfigure}
   \end{center}
   \caption{\textbf{Conceptual comparison of three masked image modeling paradigms.} 
   The difference between the three paradigms is how the parameters of the teacher network are updated.
   (a): The parameters of the teacher network are frozen during the whole training process, constructing an offline teacher.
   (b): Exponential moving average is applied to correlate the parameters of the student and teacher networks, constructing an online teacher.
   (c): \ourmethod uses a multi-stage distillation pipeline, \ie, the parameters of the teacher network are frozen except at breakpoints, where we assign student parameters to the teacher and re-initialize the student network.}
   \label{fig:conceptional difference}
\vspace{-0.2cm}
\end{figure*}

\paragraph{A random $h_\phi(\cdot)$ works surprisingly well.}
Since the choice of $h_\phi(\cdot)$ does not matter, an intuitive experiment is to see what will happen when we employ a \textbf{\textit{random teacher}}, in which the parameters are randomly initialized at the $0^{th}$ stage.
To our surprise, using a random teacher achieves performances comparably with other pre-trained teachers. 
Compared to a randomly initialized model, distilled students with multiple stages achieve 6.1\%, 20.4, and 21.3 performance gain on classification, object detection and semantic segmentation respectively. Empirically, object detection and semantic segmentation require one more stage to saturate compared to classification. 
The saturated results are on par with those induced by pre-trained teachers, which enables us to train a state-of-the-art model more efficiently, without the need of an extra pre-training stage for the initialized teacher (\eg, contrastive learning as DINO).

\section{MKD with Bootstrapped Teachers}
The study in Sec.~\ref{sec:study} motivates us to propose a multi-stage distillation pipeline for pre-training. 
The entire pre-training undergoes multiple stages split by breakpoints.
For each stage, we fix teacher network to obtain a stable visual representation, guiding the learning of student network. 
The pre-trained student model is then used as a stronger teacher and distills its knowledge to a new subsequent student, providing richer visual representations.
We re-initialize the student network at each breakpoint. 
The above process repeats itself - the teachers keep bootstrapped from the students, until a performance saturation on downstream tasks is observed. 
Hence, our strategy is to perform distillation with \textit{bootstrapped teacher}s. We illustrate our framework in \cref{fig:dbot} and the conceptual relations with the other two paradigms in~\cref{fig:conceptional difference}. 
By noting $m$ as the momentum which indicates how fast the teacher's parameters $\vtheta_t$ is updated from student's parameters $\vtheta_s$, \ie, $\vtheta_t=m\cdot \vtheta_t + (1-m)\cdot \vtheta_s$, we present the following discussions.

\paragraph{Relations with previous methods.} 
One group of works leverages \textbf{\textit{pre-trained teacher}} as in \cref{fig:pre-trainedteacher}, \ie, BEiT~\cite{beit}.
The teacher requires an extra stage of pre-training and is kept fixed with $m$ = 1. Ideally, pre-trained teachers bear additional knowledge which is prone to be more semantic meaningful, prompting student's learning. Nonetheless, the pre-training of these teachers entails a completely different computation pipeline~\cite{maskfeat} and often additional data~\cite{wei2022mvp}, complicating its practical use.
Another group as in~\cref{fig:randomteacher} works with \textbf{\textit{random teacher}} in dispense with pre-trained ones. Starting from randomness, the teachers in iBOT~\cite{zhou2021ibot} and data2vec~\cite{baevski2022data2vec}, however, are bootstrapped from the student typically with $m\in$ (0, 1), \eg, 0.9998 as in~\cite{baevski2022data2vec}. Although bootstrap induces improving quality of the teacher's representation, the pipeline is plagued by its optimization instability and sensitivity towards hyper-parameters. We note that MAE uses identity mapping of pixels as the target, which is observed to function similarly as a fixed random teacher with $m$ = 1, as shown in \cref{ap:pixel}. Despite its simplicity, such practice eludes synergy between the teacher and the student. 
 Comparatively, \ourmethod is with $m$ = 0 for every breakpoint and $m$ = 1 otherwise.

\section{Experiments}
\label{sec:exp}

\subsection{Pre-Training}

\paragraph{Architecture.}
We use different capacity Vision Transformers~\cite{vit}, \ie, ViT-B/16, ViT-L/16, and ViT-H/14 for \ourmethodn. 
The input image of size 224$\times$224 is first divided by a linear projection head into non-overlapping patch tokens total of 196 for ViT-B and ViT-L, and 256 for ViT-H. 
We exactly follow the common setup demonstrated in~\cref{sec:study}, \eg, a student with asymmetric encoder-decoder architecture, a teacher with intact input, etc.

\paragraph{Optimization.} The learning rate is first linearly increased to the initial learning rate for the first 40 epochs and then cosine annealed to 0. The initial learning rate is set as 1.5e-4 $\times$ batch\_size / 256, with batch size being 4096 for all models. We use the AdamW optimizer~\cite{adamw} and Smooth L1 loss~\cite{fast-rcnn} to optimize the parameters of student network. 
Stochastic drop rate are applied, 0.2 for ViT-B, 0.2 for ViT-L, and 0.3 for ViT-H. 
We use only center-crop and flipping for data augmentation.
As shown in \cref{tab:different teacher}, the performance of different downstream tasks saturates at different stages. 
By default, we pre-train all models for classification with 2 stages, for object detection and semantic segmentation with 3 stages.

\subsection{ImageNet Results}
\label{sec:exp-ImageNet}

We primarily focus on the end-to-end fine-tuning performance and report the top-1 validation accuracy on ImageNet-1K~\cite{imagenet} dataset.

\paragraph{Evaluation setup.}
We sweep the base learning rate within a range with a batch size being 1024. 
We warm up the learning rate during the first 5 epochs to the initial learning rate and use a cosine schedule for the rest of the epochs.
We average all the patch tokens output from the last transformer block and pass them into a linear projection head for classification.
We fine-tune ViT-B for 100 epochs and ViT-L and ViT-H for 50 epochs in total.

\begin{table}[t]
    \begin{center}
    \setlength{\tabcolsep}{2mm}
    \begin{tabular}{l|cccc}
        method& ViT-B&ViT-L&ViT-H&ViT-H$_{448}$\\
        \hline
        \textcolor{gray!80}{supervised~\cite{mae}}  & \textcolor{gray!80}{82.3}  &   \textcolor{gray!80}{82.6}    &   \textcolor{gray!80}{83.1}& \textcolor{gray!80}{-}\\
        MoCo v3~\cite{mocov3} &   83.2    &   84.1    &   -   &   -\\
        DINO~\cite{dino} &   83.6    &   -    &   -   &   -\\
        \hline
        \multicolumn{5}{l}{\textit{\small methods based on masked image modeling:}} \\
        \hline
        BEiT~\cite{beit} &   83.2    &   85.2    &   -   &   -\\
        iBOT~\cite{zhou2021ibot}    &   84.0    &   85.2    &   -   &   -\\
        MAE~\cite{mae} &   83.6    &   85.9   &  86.9  & 87.8\\
        data2vec~\cite{baevski2022data2vec}&   84.2    &   86.2    &   -   &   -\\
        \hline
        \ourmethod& \textbf{84.5}&   \textbf{86.6}&   \textbf{87.4} & \textbf{88.0}\\
    \end{tabular}
    \end{center}
    \caption{Comparison result of the previous methods on ImageNet-1K.
    We evaluate by the end-to-end fine-tuning protocol. 
    All results are based on an image size of 224, except for ViT-H with an extra result with 448 image size. 
    We perform distillation in each stage for 800 epochs and with 2 stages (our default) in total.
    }
    \label{tab:sota}
\vspace{-0.2cm}
\end{table}

\paragraph{Comparison with previous results.}
We report the fine-tuning results on ImageNet-1K, mainly focusing on the comparison of the self-supervised and supervised methods. Supervised denotes the results reported in the MAE. 
As shown in~\cref{tab:sota}, \ourmethod achieves remarkable results with different model capacities, demonstrating its scalability.
We achieved top-1 evaluation accuracy of 84.5\%, 86.6\%, and 87.4\% with ViT-B, ViT-L, and ViT-H, yielding gains of 0.9\%, 0.7\%, and 0.5\% compared to MAE. 
When fine-tuned with an image size of 448, \ourmethod further achieves an accuracy of 88.0\%, surpassing the results obtained by MAE.

\paragraph{Semi-supervised learning.}
To investigate the label efficiency of \ourmethod, we also show the semi-supervised results on ImageNet-1K under different labeled data availability in \cref{tab:semi-supervised}.
We focus on the comparison with self-supervised learning methods. 
The label-fraction sampling strategy follows~\cite{simclr}. 
\ourmethod outperforms MAE by 1.7 and 1.4 points using 1\% and 10\% of the labels, respectively, showing a higher label efficiency.

\begin{table}[t]
    \begin{center}
    \setlength{\tabcolsep}{5mm}
    \begin{tabular}{l|c|ccc}
        method & arch & 1\% & 10\% \\
        \hline
        \textcolor{gray!80}{supervised~\cite{semi-vit}} & ViT-B&\textcolor{gray!80}{-}&\textcolor{gray!80}{68.9}\\
        data2vec~\cite{baevski2022data2vec} & ViT-B&48.7& 71.2\\
        MAE~\cite{mae} & ViT-B & 53.1 & 73.1\\
        \hline
        \ourmethod & ViT-B & \textbf{54.8} & \textbf{74.5}\\ 
    \end{tabular}
    \end{center}
    \caption{Semi-supervised learning on ImageNet-1K with different self-supervised models.
    1\% and 10\% represent the label fraction.
    All results are based on our implementation with the official pre-trained model.}
    \label{tab:semi-supervised}
\vspace{-0.2cm}
\end{table}

\subsection{Downstream Tasks}
\label{sec:exp-downstream}
To further demonstrate the effectiveness, we consider dense prediction tasks: object detection, semantic segmentation, and instance segmentation, as well as classification tasks that transfer to smaller datasets.

\paragraph{Objection detection and instance segmentation.}
\begin{table}
    \begin{center}
    \setlength{\tabcolsep}{2.5mm}
    \begin{tabular}{l|cccc}
        \multirow{2}{*}{method}& \multicolumn{2}{c}{AP$^\mathrm{box}$} & \multicolumn{2}{c}{AP$^\mathrm{mask}$}\\
        &ViT-B&ViT-L&ViT-B&ViT-L\\
        \hline
        \textcolor{gray!80}{supervised~\cite{zhou2021ibot}}&\textcolor{gray!80}{49.8}&\textcolor{gray!80}{51.2}&\textcolor{gray!80}{43.2}&\textcolor{gray!80}{44.5}\\
        DINO~\cite{dino}&50.1&-&43.4&-\\
        MAE~\cite{mae}&50.6&54.0&43.9&46.2\\
        iBOT~\cite{zhou2021ibot}&51.3&-&44.3&-\\
        \hline
        \ourmethod &\textbf{52.7}&\textbf{56.0}&\textbf{45.7}&\textbf{48.2}\\
    \end{tabular}
    \end{center}
    \caption{Object detection and instance segmentation results on COCO using Cascade Mask R-CNN.
    All results are based on our implementation with the official pre-trained model.
    We perform distillation in each stage for 800 epochs and with 3 stages (default) in total. 
    }
    \label{tab:cascade det}
\vspace{-0.2cm}
\end{table}
We consider Cascade Mask R-CNN~\cite{cai2019cascade} as the task head for object detection and instance segmentation with ViT-B and ViT-L on COCO~\cite{coco}. 
We report AP$^\mathrm{box}$ and AP$^\mathrm{mask}$ for object detection and instance segmentation respectively.
The results are demonstrated in \cref{tab:cascade det}.
\ourmethod outperforms the previous self-supervised and supervised methods by a large margin, setting a new state-of-the-art result with both ViT-B and ViT-L. 
With ViT-B, \ourmethod achieves a AP$^\mathrm{box}$ of 52.7 and a AP$^\mathrm{mask}$ of 45.7, outperforming the supervised baseline pre-training by 2.9 and 2.5 points, respectively. With ViT-L, such improvement is more prominent with 4.8 and 3.6 points respectively, showing the high scalability of \ourmethod for model capacity in downstream dense prediction tasks. 

\paragraph{Semantic segmentation.}
We adapt UperNet~\cite{xiao2018unified} as the task head for semantic segmentation with ViT-B and ViT-L on ADE20K~\cite{ade}. 
We report the mIoU and mAcc for semantic segmentation, and the results are demonstrated in \cref{tab:ade20k}. 
We achieve the best performances on semantic segmentation compared to previous self-supervised methods by a nontrivial margin. 
\ourmethod improves mIoU from 47.4 to 49.5 with ViT-B, and 49.9 to 54.5 with ViT-L, yielding gains of 2.1 and 4.6 points respectively, compared to the supervised baseline.
The improvement in semantic segmentation is as significant as in object detection.

\begin{table}
    \begin{center}
    \setlength{\tabcolsep}{2.5mm}
    \begin{tabular}{l|cccc}
        \multirow{2}{*}{method}& \multicolumn{2}{c}{mIoU} & \multicolumn{2}{c}{mAcc}\\
        &ViT-B&ViT-L&ViT-B&ViT-L\\
        \hline
        \textcolor{gray!80}{supervised~\cite{mae}}&\textcolor{gray!80}{47.4}&\textcolor{gray!80}{49.9} & \textcolor{gray!80}{-} & \textcolor{gray!80}{-} \\
        iBOT~\cite{zhou2021ibot}&48.4&52.3 & 59.3 & 63.3\\
        data2vec\cite{baevski2022data2vec}&48.2&- & 59.5 &-\\
        MAE~\cite{mae}&48.1&53.6 & 58.9 & 65.5\\
        \hline
        \ourmethod&\textbf{49.5}&\textbf{54.5} & \textbf{60.7} & \textbf{66.0}\\
    \end{tabular}
    \end{center}
    \caption{Semantic segmentation results on ADE20K using UperNet. 
    All results are based on our implementation with the official pre-trained model.
    We perform distillation in each stage for 800 epochs and with 3 stages (default) in total. 
    }
    \label{tab:ade20k}
\vspace{-0.2cm}
\end{table}

\paragraph{Transfer learning.}
To further investigate the generalizability of visual representations learned by \ourmethod.
We study transfer learning performance by fine-tuning the pre-trained models on smaller datasets, including CIFAR10~\cite{krizhevsky2009learning} (Cif$_{10}$), CIFAR100~\cite{krizhevsky2009learning} (Cif$_{100}$), iNaturalist18~\cite{van2018inaturalist} (iNa$_{18}$), iNaturalist19~\cite{van2018inaturalist} (iNa$_{19}$), Flowers~\cite{nilsback2008automated} (Flwrs), and Cars~\cite{krause20133d}. 
The results are shown in \cref{tab:transfer}. \ourmethod achieves comparable, if not better, performances compared to previous best methods. 
Specifically, the improvement is significant on relatively larger datasets like iNaturalist18 and iNaturalist19, with 4.7\% and 3.3\% respectively compared to the supervised baseline.

\begin{table}
    \begin{center}
    \setlength{\tabcolsep}{0.8mm}
    \begin{tabular}{l|cccccc|c}
        {method}& Cif$_{10}$&Cif$_{100}$&iNa$_{18}$&iNa$_{19}$&Flwrs&Cars& avg. \\
        \hline
        \textcolor{gray!80}{sup.~\cite{zhou2021ibot}}& \textcolor{gray!80}{99.0}&   \textcolor{gray!80}{90.8}&   \textcolor{gray!80}{73.2}& \textcolor{gray!80}{77.7}&   \textcolor{gray!80}{98.4}&   \textcolor{gray!80}{92.1}& \textcolor{gray!80}{88.5} \\
        DINO~\cite{dino}&   99.1&   91.7&   72.6&   78.6&   98.8&   93.0&89.0 \\
        iBOT~\cite{zhou2021ibot}&   99.2&   92.2&   74.6&   79.6&   98.9&   94.3& 89.8 \\
        MAE~\cite{mae}&   -&   -& 75.4&   80.5& -& -& - \\
        \hline
        \ourmethod&\textbf{99.3}&91.3&\textbf{77.9}&\textbf{81.0}&98.2&93.7&\textbf{90.2} \\
    \end{tabular}
    \end{center}
    \caption{Transfer classification accuracy on various datasets. We report the results of ViT-B.
    Sup. denotes the supervised baseline. The average results (avg.) are shown in the rightmost column. 
    }
    \label{tab:transfer}
\vspace{-0.2cm}
\end{table}

\begin{table*}[!ht]
\vspace{-.2em}
\begin{center}
\subfloat[
\textbf{Stage split number}.
2-stage distillation works the best.
\label{tab:split number}
]{
\begin{minipage}{0.29\linewidth}{\begin{center}
\tablestyle{1pt}{1.05}
\begin{tabular}{y{64}x{24}}
pre-training epochs & acc \\
\shline
1600 & 83.6 \\
800-800    &   \baseline{\textbf{84.5}}\\
533-533-533 &   84.4\\
\\
\end{tabular}
\end{center}}\end{minipage}
}
\hspace{2em}
\subfloat[
\textbf{Epoch for each stage}.
2-stage distillation with 800 epochs for each stage works the best.
\label{tab:each stage}
]{
\begin{minipage}{0.29\linewidth}{\begin{center}
\tablestyle{6pt}{1.05}
\begin{tabular}{y{64}x{24}}
pre-training epochs & acc \\
\shline
400-800 &  84.3 \\
800-400 &  84.3 \\
800-800 & \baseline{\textbf{84.5}} \\
800-1200 &  84.3 \\
\end{tabular}
\end{center}}\end{minipage}
}
\hspace{2em}
\subfloat[
\textbf{Momentum update}.
The \textit{vanilla} strategy explicitly splitting stages works the best.
\label{tab:momentum}
]{
\begin{minipage}{0.29\linewidth}{\begin{center}
\tablestyle{1pt}{1.05}
\begin{tabular}{y{64}x{24}}
momentum & acc \\
\shline
\textit{vanilla} & \baseline{\textbf{84.5}}\\
0.9998 & 83.6\\
0.9999 &  83.9 \\
cosine(0.996,1) & 82.1\\
\end{tabular}
\end{center}}\end{minipage}
}
\\
\vspace{.4em}
\subfloat[
\textbf{Target normalization}.
Using patch representations w/o \texttt{[LN]} as targets works best.
\label{tab:reconstruct target}
]{
\begin{minipage}{0.29\linewidth}{\begin{center}
\tablestyle{4pt}{1.05}
\begin{tabular}{y{64}x{24}}
target norm & acc \\
\shline
w/ \texttt{[LN]} & 84.3 \\
w/o \texttt{[LN]} & \baseline{\textbf{84.5}}\\
\\
\end{tabular}
\end{center}}\end{minipage}
}
\hspace{2em}
\subfloat[
\textbf{Student initialization}.
Re-initializing the student's weight at breakpoints works best.
\label{tab:student weight}
]{
\centering
\begin{minipage}{0.29\linewidth}{\begin{center}
\tablestyle{4pt}{1.05}
\begin{tabular}{y{64}x{24}}
student init & acc \\
\shline
w/o re-initialize & 84.2 \\
w/ re-initialize & \baseline{\textbf{84.5}} \\
\\
\end{tabular}
\end{center}}\end{minipage}
}
\centering
\hspace{2em}
\subfloat[
\textbf{Mask ratio}. 
A mask ratio of 75\%  works best.
\label{tab:mask ratio}
]{
\centering
\begin{minipage}{0.29\linewidth}{\begin{center}
\tablestyle{4pt}{1.05}
\begin{tabular}{y{64}x{24}}
mask ratio & acc \\
\shline 
0.7 & 84.3 \\
0.75 & \baseline{\textbf{84.5}} \\
0.8 & 84.2 \\
\end{tabular}
\end{center}}\end{minipage}
}
\vspace{-.4em}
\end{center}
\caption{\textbf{Ablation study} with \textbf{ViT-B/16} on ImageNet-1K validation set. We report with the end-to-end fine-tuning top-1 accuracy (\%).
Ablation study is conducted with randomly initialized teachers. We note that models distilled from the pre-trained teachers generally share similar trends.
Default settings are marked in \colorbox{baselinecolor}{gray}.
\textit{vanilla} denotes $m$ being 0 at the breakpoint and 1 otherwise.
cosine(a,b) denotes $m$ is cosine annealed from value a to b.
}
\label{tab:ablations} \vspace{-.5em}
\end{table*}

\subsection{Ablation Study}

\paragraph{Stage split number.}
We study the influence of stage number by splitting total training epochs of 1600 into varying distillation stages, from 0 to 2.
Results are shown in \cref{tab:split number}. 
2-stage distillation works the best (for classification task), achieving 84.5\% accuracy.
Splitting epochs to 3-stage brings 0.1\% performance drop,
while all splitting strategies obtain a top-1 accuracy higher than 83.6\%, indicating its generalizability.

\paragraph{Epoch for each stage.}
\cref{tab:each stage} studies proper epochs needed for each stage in a 2-stage distillation pipeline. 
With the 2$^{\textrm{nd}}$ stage distilling for 800 epochs, longer epochs for the 1$^{\textrm{st}}$ stage induces 0.2\% improvement (84.3\% \textit{vs.} 84.5\%).
With the 1$^{\textrm{st}}$ stage distilling for 800 epochs, 800 epochs are enough for the 2$^{\textrm{nd}}$ stage since 1200 epochs incur no gain.
Evenly splitting the epochs in 2-stage masked knowledge distillation achieves the best performance.

\paragraph{Momentum update.}
We use in \ourmethod a multi-stage distillation pipeline, which is to distill from a momentum encoder with $m$ being 0 for every breakpoint and 1 otherwise. 
We further investigate other momentum update strategies commonly used in self-supervised learning.
Results are shown in \cref{tab:momentum}.
The \textit{vanilla} strategy works the best.

\paragraph{Target normalization.}
We study whether patch tokens obtained by the self-attention blocks to be used as target representation should be passed through the Layer Normalization~\cite{layernorm} layer \texttt{[LN]}.
The accuracy of models after 2-stage distillation is shown in \cref{tab:reconstruct target}.
Without passing through \texttt{[LN]}, the patch tokens directly obtained from the transformer block make them less suitable as target representations to guide students' learning.

\paragraph{Student initialization.}
We study whether student's weight should remain when entering the next stage of distillation. Specifically, we either keep the student's weight unchanged or re-initialize the student at each breakpoint.
As shown in \cref{tab:student weight}, re-initializing the student's weight works the best.

\paragraph{Mask ratio.}
\cref{tab:mask ratio} shows the influence of the mask ratio on end-to-end fine-tuning.
The optimal mask ratio for \ourmethod is 75\%, the same as that in MAE.

\begin{figure*}[ht]
    \centering
    \includegraphics[width=1.0\linewidth]{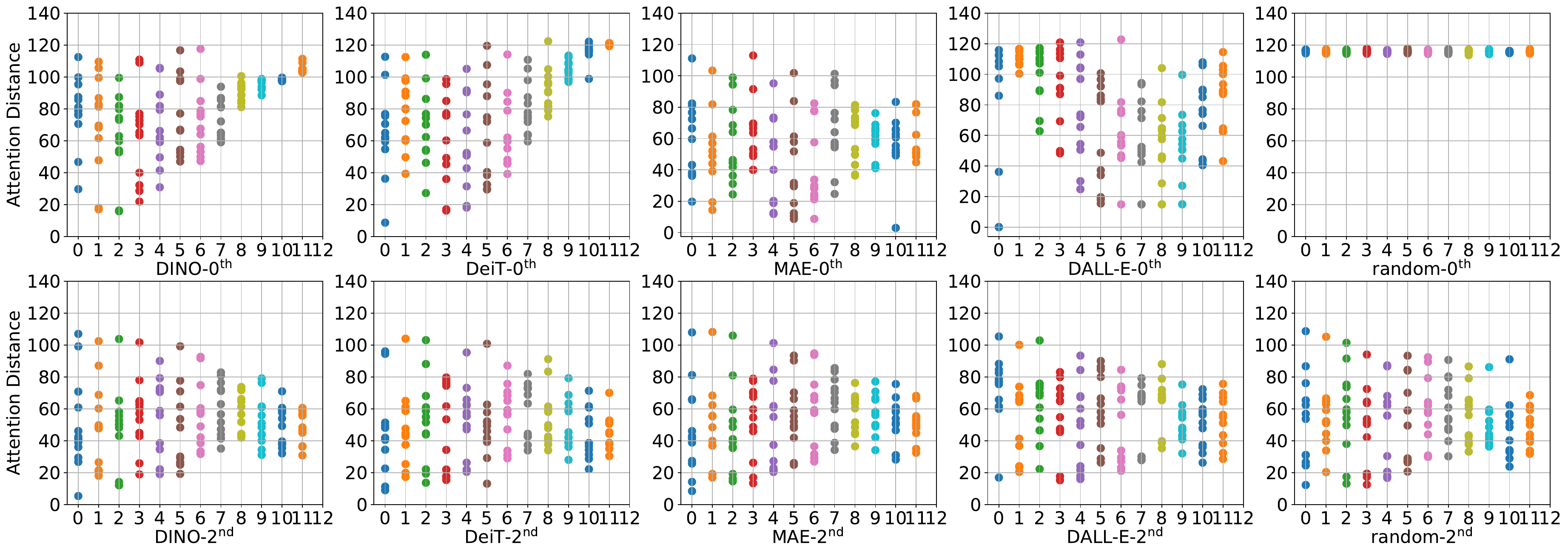}
    \caption{Average attention distance of different heads w.r.t layer number of ViT-B with different distilling teachers and their corresponding student distilled for 2 stages. The first row showcases the teachers while the second showcases the 2$^\mathrm{th}$ stage distilled student. Models using different teachers achieve the same result. The distilled students obtain comparatively more local attention compared to the teachers.}
    \label{fig:avd}
\end{figure*}

\section{Property Analysis}
\label{sec:analysis}
We investigate the properties of models distilled from different teachers under certain criteria, analyzing models' weights and outputs. Further, training efficiency is briefly discussed with previous methods.

\paragraph{Averaged attention distance.}
We compute averaged attention distance~\cite{vit}, averaged over ImageNet-1K val set, for each attention head of different blocks to understand how local and global information flows into Transformers. 
Average attention distance for \ourmethod using DeiT, DINO, MAE, DALL-E, and random as teachers are illustrated in \cref{fig:avd}.
The higher the attention distance, models' attention over an image is more global.
Although the average attention distance of disparate initialized teachers varies greatly, their distilled students after multi-stage distillation exhibit similar behaviors, \eg, models' attention toward local or global contents.
Additionally, \ourmethod achieves more local attention than previous works.

\begin{figure*}[ht]
  \centering
  \includegraphics[width=1.0\linewidth]{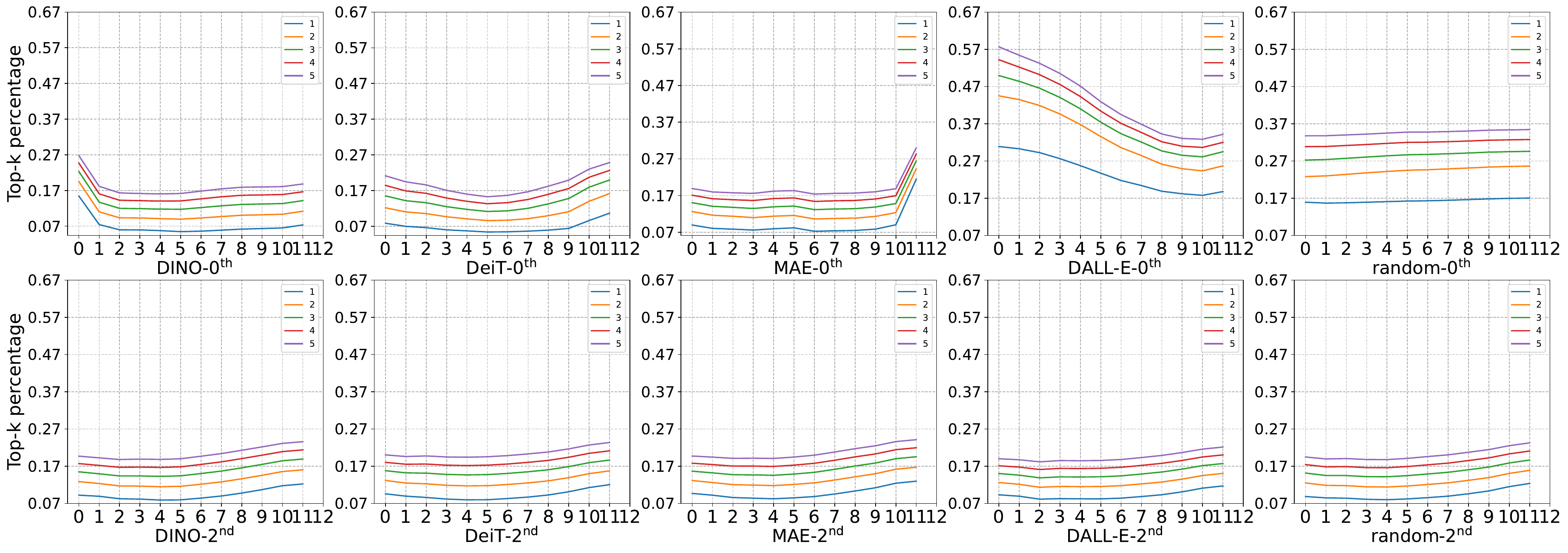}
  \caption{Singular value decomposition of different layers of ViT-B with different distilling teachers and their corresponding student distilled for 2 stages. The first row showcases the teachers while the second showcases the 2$^\mathrm{th}$ stage distilled student. Models using different teachers achieve the same result.}
  \label{fig:svd}
\end{figure*}


\paragraph{Singular value decomposition.}
We computed the percentage of top-$k$ singular values~\cite{svd} of the embedding w.r.t each layer. The results are averaged over the ImageNet-1K val set. We showcase the results with $k$ varying from 1 to 5. 
Singular value decomposition for \ourmethod using DeiT, DINO, MAE, DALL-E, and random as teachers are shown in \cref{fig:svd}.
The higher the percentage, the models' output over an image is less correlated, indicating larger redundancy of its spatial representations thus less suitability for compression. Intuitively, random models at the 0$^\textrm{th}$ stage has the largest percentage given that pixel are merely randomly projected.
The student networks distilled from different initialized teachers exhibit similar behaviors.

\begin{table}[t]
	\begin{center}
        \begin{tabular}{l|ccccc}
            model & DeiT & DINO & DALL-E & MAE & random \\
            \hline
            0$^\mathrm{th}$ & 13.8 & 36.1 & 36.5 & 36.6 & 23.6 \\
            2$^\mathrm{nd}$ & 36.6 & 36.6 & 36.6 & 36.6 &36.6\\
        \end{tabular}
        \end{center}
    \caption{The results of unsupervised object detection on Pascal VOC 2012 with CorLoc based on SVD decomposition.}
    \label{tab:unsup-det}
\vspace{-0.4cm}
\end{table}

\paragraph{Unsupervised object detection.}
We use unsupervised object localization to quantitatively evaluate the visual representation obtained by different models.
We follow the evaluation practice proposed in~\cite{unsup-det} with Correct Localization (CorLoc) on POC-VOC 2012 trainval sets, except that we conduct feature decomposition via SVD instead of Laplacian since we observe more stable behaviors with SVD.
We first compute singular value decomposition for the patch feature obtained by the ViT-B last block. Then a sign operation is applied on the first eigenvector, obtaining a binary mask of an image. We then take the bounding box around the largest connected component, which is more like the foreground object instead of the background. Correct localization (CorLoc) is used to measure the results, evaluated on POC-VOC 2012 trainval sets. A box is considered to have correctly identified an object if it has more than 50\% intersection-over-union with a ground truth bounding box.
Quantitative results are demonstrated in \cref{tab:unsup-det}. \ourmethod using different teachers achieves very similar results, with students consistently outperforming their teachers.

\begin{table}{}
    \setlength{\tabcolsep}{1.4mm}
	\begin{center}
        \begin{tabular}{l|cccc}
            method & data2vec~\cite{baevski2022data2vec} & BEiT~\cite{beit} & MAE~\cite{mae} &{\ourmethod} \\
            \hline
            \textit{asym.} & \xmark & \xmark & \cmark  & \cmark \\
            \hline
            ViT-B & 169 & 166 & 79   & 109 \\
            ViT-L & 431 & 356 & 125 & 200 \\
            ViT-H & 960 & 751 & 240 &  416 \\
        \end{tabular}
        \end{center}
    \caption{Training time (s) per epoch for different methods with ViT-B/16, ViT-L/16, and ViT-H/14. \textit{asym.} denotes whether to use an asymmetric encoder-decoder structure~\cite{mae}. All entries are tested on the same setting, \ie, with 32 NVIDIA A100-80G GPUs.}
    \label{tab:time}
\vspace{-0.4cm}
\end{table}

\paragraph{Training efficiency.}
We compute the training time per epoch for different methods in \cref{tab:time}. 
With an asymmetric encoder-decoder architecture (\textit{asym.}) as the default setup, \ourmethod performs slower than MAE, but much faster than data2vec and BEiT. Such advantage turns more significant with models of larger size.

\section{Distill from Bigger Teachers}
\label{sec:biggert}

Inspired by canonical practices in knowledge distillation~\cite{kd}, we use larger teachers to distill smaller students, showcasing the potential of MKD in general. 
Specifically, we attempt to use ViT-L/H as teacher networks to distill ViT-B, and ViT-H as the teacher network to distill ViT-L. All larger teachers are first distilled for 2 stages with the default setup.
We resize the image to 196$\times$196 for ViT-H/14 to keep the length of its output the same as that of ViT-B/L.
While we do not find substantial gains on classification results, the results by distilling from ViT-H are significantly better for dense prediction tasks compared to the default setup, \ie, +0.8 points of AP$^\mathrm{box}$ and +1.3 points of mIoU with ViT-B as the student. 
The performance gain in distilling ViT-L from ViT-H is diminished but still valid, \ie, +0.1 AP$^\mathrm{box}$ and +0.7 mIoU.
We also consider MKD with data-richer teachers, \eg CLIP, as exploratory experiments and set new state-of-the-art results for self-supervised learning. Refer to~\cref{app:sec:datarichert} for details.

\begin{table}
    \begin{center}
    \setlength{\tabcolsep}{1.2mm}
    \begin{tabular}{l|c|ccc}
        teacher& student& cls. & det.& seg. \\
        \hline
        \textcolor{gray!80}{ViT-B}&\multirow{3}{*}{ViT-B} & \textcolor{gray!80}{84.5} \hspace{0.88cm} & \textcolor{gray!80}{52.7} \hspace{0.88cm} & \textcolor{gray!80}{49.5} \hspace{0.88cm} \\
        ViT-L& & 84.6 \textcolor{mgreen}{(+0.1)}&   53.1 \textcolor{mgreen}{(+0.4)}&   50.1 \textcolor{mgreen}{(+0.6)}\\
        ViT-H& &   84.6 \textcolor{mgreen}{(+0.1)}&   53.5 \textcolor{mgreen}{(+0.8)}&   50.8 \textcolor{mgreen}{(+1.3)}\\
        \hline
        \textcolor{gray!80}{ViT-L} & \multirow{2}{*}{ViT-L} &\textcolor{gray!80}{86.6} \hspace{0.88cm} & \textcolor{gray!80}{56.0} \hspace{0.88cm} &\textcolor{gray!80}{54.5} \hspace{0.88cm} \\
        ViT-H && 86.8 \textcolor{mgreen}{(+0.2)} & 56.1 \textcolor{mgreen}{(+0.1)} & 55.2 \textcolor{mgreen}{(+0.7)}\\
    \end{tabular}
    \end{center}
    \caption{Results of classification (cls.) on IN1K, object detection (det.) on COCO, and semantic segmentation (seg.) on ADE20K. 
    For same-size teachers (colored \textcolor{gray!80}{gray}), students are pre-trained with default settings.
    For bigger teachers, students are pre-trained for 1-stage from 2-stage distilled teachers.}
    \label{tab:big model}
\vspace{-0.3cm}
\end{table}

\section{Conclusion}
As a special case of MIM, we formulate MKD upon which an empirical investigation is conducted about the influence of different target representations on self-supervised masked autoencoders. 
The study concludes that it is \textit{not necessary} to carefully choose the target representation to learn good visual representations if distillation is performed in multiple stages (\ie, with bootstrapped teachers).
Instead of initializing teachers with pre-trained models, we resort to random ones for simple practice.
\textit{Without an extra stage of pre-training}, \ourmethod achieves favorable performance on image classification, object detection, and semantic segmentation.
We hope our study and method will provide timely insights for self-supervised learning.

{\small
\bibliographystyle{ieee_fullname}
\bibliography{egbib}
}


\clearpage
\newpage
\appendix

\renewcommand\thesection{\Alph{section}} 
\renewcommand\thesubsection{\Alph{section}.\arabic{subsection}} 
\renewcommand\thefigure{\Alph{section}\arabic{figure}} 
\renewcommand\thetable{\Alph{section}\arabic{table}} 
\setcounter{section}{0}
\setcounter{figure}{0}	
\setcounter{table}{0}

\section{Implementation Details}
\label{app:implementation detail}

\subsection{Pre-Training}
\label{app:pretrain}

\paragraph{Default setup.} We show our \textit{default} pre-training setup in the second colum of~\cref{tab:app:presetting}.
We use Xavier Uniform~\cite{xinitialize} to initialize the Vision Transformer~\cite{vit}. Note that we use asymmetry stochastic drop path rate for students and teachers.

\paragraph{Setup for distillation from bigger teachers.} We follow the \textit{default} setup, except that we use a different setup for stages. We first train larger-size teachers for 2 stages (in all downstream tasks) and use those to distill new students for 1 stage (in all downstream tasks).

\subsection{Classification}
\label{app:imagenetcls}

The \textit{default} end-to-end fine-tuning recipe is shown in the second column of~\cref{tab:app:fine-tune}, following the common recipes~\cite{mae,beit} of ViT tuning for self-supervised models. The same recipe is applied when distilling from bigger teachers.

\subsection{Object Detection and Instance Segmentation}
\label{ap:det}
We adopt the vanilla ViT with Cascade Mask R-CNN~\cite{cai2019cascade} as the task head on COCO~\cite{coco} dataset for object detection and instance segmentation, following the common setup~\cite{zhou2021ibot}.
The default recipe is shown in \cref{tab:app:det-iseg}. To cope with versatile image sizes, we add relative position embedding instead of interpolating the absolute position embedding obtained during pre-training.
For a fair comparison, we applied the same setup and sweep the learning rate and stochastic drop path rate for different methods.

\subsection{Semantic Segmentation}
\label{ap:seg}
We use vanilla ViT and UperNet~\cite{xiao2018unified} as the task head on ADE20K~\cite{ade} dataset for semantic segmentation, following the common setup~\cite{beit}.
The default recipe is shown in \cref{tab:app:seg}. To cope with versatile image sizes, we add relative position embedding instead of interpolating the absolute position embedding obtained during pre-training.
For a fair comparison, we applied the same setup and sweep the learning rate and layer-wise decay for different methods.

\begin{table}[t]
    \begin{center}
    \setlength{\tabcolsep}{0.5mm}
	\begin{tabular}{c|cc}
	\toprule
		config & default & \hspace{0.6cm} recipe\alambic \hspace{0.6cm} \\
		\hline
		optimizer & \multicolumn{2}{c}{AdamW ~\cite{adamw}} \\
		optim. momentum $\beta_1$ & \multicolumn{2}{c}{0.9} \\
		optim. momentum $\beta_2$ & 0.95 & 0.98 \\
		loss & Smooth L1 & negative cos. \\
		peak learning rate & 2.4e-3 & 3e-3 \\
		learning rate schedule & \multicolumn{2}{c}{cosine decay ~\cite{cosinedecay}} \\
		batch size & \multicolumn{2}{c}{4096} \\
		weight decay & \multicolumn{2}{c}{0.05} \\
		stages & 2 (c.), 3 (d./s.)  & 1 \\
		epochs per stage & 800  & 1600 \\
		warmup epochs~\cite{warmup} & 40  & 10 \\
		augmentation & \multicolumn{2}{c}{RandomResizedCrop} \\
		aug. input scale & (0.2, 1) & (0.4, 1) \\
		asym. enc-dec~\cite{mae} & \cmark  & \xmark\\
		drop path~\cite{droppath} & 0.2 (B/L), 0.3 (H) & 0.1 (B/L/H) \\
		target w/ \texttt{[LN]} & \xmark & \cmark \\
		mask ratio & 0.75  & 0.4 \\
	\bottomrule
	\end{tabular}
        \end{center}
	\caption{\textbf{Pre-training setup.} recipe\alambic is the pre-training recipe for \ourmethoddis. cos. denotes cosine distance. c., d., and s. denotes downstream tasks of classification, object detection, and semantic segmentation respectively. drop path is for the students.}
	\label{tab:app:presetting}
\end{table}

\begin{table}[t]
    \begin{center}
    \setlength{\tabcolsep}{0.2mm}
	\begin{tabular}{c|cc}
	\toprule
		config & \hspace{0.8cm}default & \hspace{0.8cm} recipe\alambic \hspace{0.8cm} \\
        \hline
		optimizer & \multicolumn{2}{c}{AdamW~\cite{adamw}} \\
        peak learning rate & \{0.8,1.2,1.6,2\}e-3 & \{1,2,3,4\}e-4 \\
		weight decay & \multicolumn{2}{c}{0.05} \\
		optim. momentum & \multicolumn{2}{c}{$\beta_1, \beta_2 = 0.9, 0.999$} \\
		layer-wise decay & \multicolumn{2}{c}{0.75} \\
		batch size & \multicolumn{2}{c}{1024} \\
		learning schedule & \multicolumn{2}{c}{cosine decay} \\
		warmup epochs & \multicolumn{2}{c}{5} \\
		epochs & \multicolumn{2}{c}{100 (B), 50 (L/H)} \\
		augmentation & \multicolumn{2}{c}{RandAug (9, 0.5) ~\cite{randaug}} \\
		label smoothing & \multicolumn{2}{c}{0.1} \\
		mixup~\cite{mixup} & \multicolumn{2}{c}{0.8} \\
		cutmix~\cite{cutmix} & \multicolumn{2}{c}{1.0} \\
		drop path~\cite{droppath} & 0.2 (B/L), 0.3 (H) & 0.1 (B), 0.2 (L), 0.3 (H) \\
		\bottomrule
	\end{tabular}
 \end{center}
	\caption{\textbf{End-to-end fine-tuning setup.} recipe\alambic is the pre-training recipe for \ourmethoddis.}
	\label{tab:app:fine-tune}
\end{table}

\begin{table}[t]
    \begin{center}
    \setlength{\tabcolsep}{7.0mm}
	\begin{tabular}{c|c}
	\toprule
		config & value \\
		\hline
		optimizer & AdamW~\cite{adamw}\\
		optim. momentum & $\beta_1, \beta_2 = 0.9, 0.999$ \\
		peak learning rate & 1e-4 \\
		batch size & 16 \\
		layer-wise decay & 0.75 \\
		weight decay & 0.05 \\
		learning schedule & step \\
		epochs & 12 \\
		step epochs & 8, 11 \\
		drop path~\cite{droppath} & 0.2 \\
	\bottomrule
	\end{tabular}
 \end{center}
	\caption{\textbf{Object detection and instance segmentation setup.}}
	\label{tab:app:det-iseg}
\end{table}

\begin{table}[t]
    \begin{center}
    \setlength{\tabcolsep}{6.0mm}
	\begin{tabular}{c|c}
	\toprule
	    config & value \\
		\hline
		optimizer & AdamW~\cite{adamw}\\
		optim. momentum & $\beta_1, \beta_2 = 0.9, 0.999$ \\
		peak learning rate & \{0.3,0.5,0.8,1,3\}e-4 \\
		batch size & 16 \\
		layer-wise decay & \{0.65,0.75,0.85.0.95\}\\
		weight decay & 0.05\\
		learning schedule & cosine \\
		steps & 16000 \\
		warmup steps & 1500 \\
		drop path~\cite{droppath} & 0.1(B), 0.2(L) \\
	\bottomrule
	\end{tabular}
 \end{center}
	\caption{\textbf{Semantic segmentation setup.}}
	\label{tab:app:seg}
\end{table}

\section{Additional Experiments}

\subsection{Pixels \textit{vs.} Random Mapping of Pixels}
\label{ap:pixel}
MAE performs masked image modeling using the image pixel as the reconstruction target. 
We directly alter the target to patch tokens obtained from the image fed into a randomly initialized network. 
We select two patch tokens as the reconstruction target, one is the token obtained using the last transformer block, and the other is the token obtained using linear projection, \ie, without any transformer block.
After 400 epoch pre-training of ViT-B, the top-1 accuracy of the model on ImageNet-1K obtained by the three different targets is shown below.
\begin{table}[h]
    \vspace{-0.5em}
    \begin{center}
    \setlength{\tabcolsep}{4.5mm}
    \begin{tabular}{cccc}
        epoch & pixel & 0$^\mathrm{th}$ block & 12$^\mathrm{th}$ block \\
        \hline
        400 & 83.3 & 83.2 & 83.2 \\
        1600 & 83.6 & 83.6 & 83.6
    \end{tabular}
    \end{center}
\vspace{-0.7cm}
\end{table}

It can be derived that using the patch token obtained by a randomly initialized network as the target can achieve comparable results with a pixel as a target. 
A similar result proves that patch tokens obtained by a randomly initialized can also serve as a good reconstruction target.

\subsection{Object Detection with Mask R-CNN}
\label{ap: maskrcnn coco}
Additionally, we use Mask R-CNN structure with FPN for object detection and instance segmentation on COCO datasets. 
The results are shown in \cref{app:tab:cocodet}. \ourmethod outperforms other methods by a large margin, which is similar to the results using Cascade Mask R-CNN.

\subsection{Linear Probing}
We evaluate the linear probing performance of \ourmethod and MAE using ViT-B following the same setup as MAE, the results of which is shown below. 
\begin{table}[h]
\vspace{-0.2cm}
    \begin{center}
    \setlength{\tabcolsep}{3mm}
    \begin{tabular}{cc}
        MAE & dBOT  \\
        \hline
        67.8\% & 67.9\% \\
    \end{tabular}
    \end{center}
    \label{tab:my_label}
\vspace{-0.8cm}
\end{table}

dBOT achieves comparable linear probing performances with MAE.

\begin{table}[t]
    \begin{center}
    \setlength{\tabcolsep}{3mm}
    \begin{tabular}{l|cccc}
        \multirow{2}{*}{method}& \multicolumn{2}{c}{AP$^\mathrm{box}$} & \multicolumn{2}{c}{AP$^\mathrm{mask}$}\\
        &ViT-B&ViT-L&ViT-B&ViT-L\\
        \hline
        \textcolor{gray!80}{supervised}&\textcolor{gray!80}{47.9}&\textcolor{gray!80}{49.3}&\textcolor{gray!80}{42.9}&\textcolor{gray!80}{43.9}\\
        iBOT~\cite{zhou2021ibot}&48.6&50.6&43.1&44.7\\
        data2vec~\cite{baevski2022data2vec}&41.1&46.1&37.0&41.0\\
        MAE~\cite{mae}&50.2&53.5&44.8&47.4\\
        \hline
        \ourmethod&\textbf{51.4}&\textbf{54.0}&\textbf{45.8}&\textbf{48.0}\\
    \end{tabular}
    \end{center}
    \caption{Object detection and instance segmentation results on COCO using \textbf{Mask R-CNN}. We report the result both with ViT-B and ViT-H. All results are based on our implementation with official released pre-trained model.}
    \label{app:tab:cocodet}
\end{table}

\section{Distill from Data-Richer Teachers}
\label{app:sec:datarichert}

\begin{table}[t]
    \begin{center}
    \setlength{\tabcolsep}{2.8mm}
    \begin{tabular}{l|c|c|c}
        \makecell[l]{initialized\\ teacher} &pre-training data &\makecell[c]{asym. \\enc-dec} & acc \\
        \hline
        \textcolor{gray!80}{random}&\textcolor{gray!80}{IN1K} &\textcolor{gray!80}{\cmark}  & \textcolor{gray!80}{84.5} \\
        random&IN1K &\xmark  & 83.8\\
        \hline
        \textcolor{gray!80}{DINO~\cite{dino}}&\textcolor{gray!80}{IN1K} &\textcolor{gray!80}{\cmark} & \textcolor{gray!80}{84.4} \\
        DINO~\cite{dino}&IN1K &\xmark & 84.8 \\
        CLIP~\cite{clip}&IN1K + 400M ITp.&\cmark  & 84.9\\
        CLIP~\cite{clip}&IN1K + 400M ITp.&\xmark  &85.7 \\
    \end{tabular}
    \end{center}
    \caption{
    Image classification on IN1K with DINO and CLIP as initialized teachers, as well as random ones. Students with DINO and CLIP as teachers are distilled for 1 stage.}
    \label{tab:multi-modal model}
\vspace{-0.2cm}
\end{table}

\begin{table}[t]
	\begin{center}
	    \setlength{\tabcolsep}{1.0mm}
        \begin{tabular}{l|ccccc|c}
            \makecell[l]{pre-training\\epochs} & \makecell[c]{ran-\\dom} & \makecell[c]{DALL-E\\\cite{dall-e}} & \makecell[c]{DeiT\\\cite{deit}} & \makecell[c]{DINO\\\cite{dino}} & \makecell[c]{MAE\\\cite{mae}} & \makecell[c]{CLIP\\\cite{clip}} \\
            \hline
            0 & 77.3 & 81.1 & 81.8 & 83.2 & 83.6 & 84.8 \\
            1$\times$1600 & 83.6 & 83.6 & 83.6 & 84.4  & 84.4 & 84.9 \\
            2$\times$800 & 84.5 & 84.4 & 84.3 & 84.5 & 84.4 & 84.0 \\
            \hline
            $\triangle$ & \textcolor{mgreen}{+0.9} & \textcolor{mgreen}{+0.8} & \textcolor{mgreen}{+0.7} & \underline{\textcolor{mgreen}{+0.1}} &  \underline{\textcolor{mgreen}{+0.0}} & \underline{\textcolor{red}{-0.9}}\\
        \end{tabular}
        \end{center}
    \caption{ImageNet-1K classification results of 1 stage masked knowledge distillation with different teachers. Total epochs are shown in the format of (stages$\times$epochs\_per\_stage). $\triangle$ denotes performance gaps between entries of 2$\times$800 and 1$\times$1600. 
    }
    \label{tab:app:one stage}
\end{table}

\begin{table}[t]
    \begin{center}
    \setlength{\tabcolsep}{0.7mm}
    \begin{tabular}{l|c|ccc}
        \makecell[l]{initialized\\ teacher} &student & cls.& det. & seg. \\
        \hline
        \textcolor{gray!80}{random} & \multirow{2}{*}{ViT-B} & \textcolor{gray!80}{84.5} \hspace{0.88cm} & \textcolor{gray!80}{52.7} \hspace{0.88cm} & \textcolor{gray!80}{49.5} \hspace{0.88cm} \\
        CLIP-B~\cite{clip} & & 85.7 \textcolor{mgreen}{(+1.2)} &
        53.6 \textcolor{mgreen}{(+0.9)} &
        52.9 \textcolor{mgreen}{(+3.4)}  \\
        \hline
        \textcolor{gray!80}{random} & \multirow{2}{*}{ViT-L} &\textcolor{gray!80}{86.6} \hspace{0.88cm} & 
        \textcolor{gray!80}{56.0}\hspace{0.88cm} &
        \textcolor{gray!80}{54.5} \hspace{0.88cm} \\
        CLIP-L~\cite{clip} & &
        87.8 \textcolor{mgreen}{(+1.2)} & 
        56.8 \textcolor{mgreen}{(+0.8)}&
        56.2 \textcolor{mgreen}{(+1.7)}\\
        \hline
        \textcolor{gray!80}{random} & \multirow{2}{*}{ViT-H} &\textcolor{gray!80}{87.4} \hspace{0.88cm} & \textcolor{gray!80}{-}&\textcolor{gray!80}{-} \\
        CLIP-L~\cite{clip}& &
        88.5 \textcolor{mgreen}{(+1.1)} &- & -\\
    \end{tabular}
    \end{center}
    \caption{Results of classification (cls.) on IN1K, object detection (det.) on COCO, and semantic segmentation (seg.) on ADE20K with CLIP~\cite{clip} as the teacher. Students are distilled for 1 stage. The det. results with CLIP as teachers are with absolute positional embedding.}
    \label{tab:clip-downstream}
\vspace{-0.4cm}
\end{table}

We explore to use models pre-trained with richer data (\ie, CLIP~\cite{clip} with 400M Image-Text pairs) as the initialized teacher to seek a potential upper-bound of MKD. 

\subsection{Pre-Training}
Compared to the \textit{default} setup, there exist two major disparities of the pre-training recipes for models distilled from data-richer teachers, discussed next. The following practice is summarized as \textit{recipe\alambic} detailed in~\cref{tab:app:presetting}. 

\noindent\textbf{Vanilla Architecture.} We find that not using the asymmetric encoder-decoder architecture~\cite{mae} is optimal, as shown in~\cref{tab:multi-modal model}. While an asymmetric architecture generates momentum for bootstrapping models similar to~\cite{byol}, which lies crucial for distillation with random teachers, it hurts the performance when distilling with stronger pre-trained teachers. 

Hypothetically, the significance of the decoder in asymmetrical encoder-decoder architecture lies in the need for separate layers to decode low-level details when the targets contain little semantics (\eg, pixels and random mappings of pixels). 
Such a need is eased when the target contains high-level semantics (\eg, DINO and CLIP). The existence of the decoder, in this case, may even restrain the encoder to grasp full knowledge from the teacher, inducing degraded performances.

\noindent\textbf{1-Stage MKD}. 
We use different models as teachers to distill students for one stage with longer epochs, \ie, 1600. Results are shown in \cref{tab:app:one stage}.
Empirically, the performance gains for multi-stage MKD over 1-stage MKD decrease as teachers' fine-tuning performance increases. Stronger teachers, such as DINO and MAE, induce similarly performed students with 1-stage MKD (1$\times$1600) compared to 2-stage MKD (2$\times$800). 

Specifically, when using CLIP as the pre-trained teacher, the performance for 2-stage MKD is, to our surprise, 0.9\% lower than that of 1-stage MKD. Understandably, although the fine-tuning result of the student after 1-stage distillation is better than that of CLIP, the student is essentially trained on IN1K and may not contain faithfully data information stored in the CLIP model. Therefore, strong teachers work well with 1-stage MKD, especially for models pre-trained on extra richer data.

\subsection{Downstream Tasks}

\noindent\textbf{Implementation Details.} For fine-tuning, we also use a slightly different recipe from \textit{default} one with smaller learning rates and drop path, dubbed as \textit{recipe\alambic} detailed in~\cref{tab:app:fine-tune}. For object detection, instance segmentation, and semantic segmentation, we follow the default setup detailed in~\cref{ap:det,ap:seg}.

\noindent\textbf{Results.} Results for downstream tasks are shown in \cref{tab:clip-downstream}. 
ViT-B distilled from CLIP-B achieves an 85.7\% top-1 accuracy and a 52.9 mIoU, surpassing all previous arts.
With CLIP-L as the teacher, ViT-H with image resolution $448$ achieves an \textbf{89.1\%} top-1 accuracy, setting a new state-of-the-art image recognition result.

\subsection{Conflict with Main Conclusion}

It can be observed that MKD with CLIP~\cite{clip} as the teacher performs much better than that with the random teacher and multi-stage distillation, which seems contradictory to our main conclusion that \textit{teacher networks do not matter with multi-stage masked knowledge distillation.}
Notably, CLIP is trained with 400M image text pairs (300$\x$ larger than ImageNet-1K), which is a drastically different setup from multi-stage distillation on \underline{ImageNet-1K only}. 
Exploring CLIP as a target representation gains popularity~\cite{wei2022mvp} recently but is beyond the main scope of this paper. We present these results to corroborate the validity and to explore the upper bound of MKD in general.
We note that the exact solution to resolve the conflict is to perform multi-stage distillation using the CLIP's in-house 400M data to which we have no access. It is hypothesized that two results should be matched in light of experiments on  ImageNet-1K, which is left to future work.

\end{document}